\documentclass{article}

\usepackage[preprint]{neurips}
\PassOptionsToPackage{numbers, compress}{natbib}

\usepackage[utf8]{inputenc} % allow utf-8 input
\usepackage[T1]{fontenc}    % use 8-bit T1 fonts
\usepackage{hyperref}       % hyperlinks
\usepackage{url}            % simple URL typesetting
\usepackage{booktabs}       % professional-quality tables
\usepackage{amsfonts}       % blackboard math symbols
\usepackage{nicefrac}       % compact symbols for 1/2, etc.
\usepackage{microtype}      % microtypography
\usepackage{amsmath,amsthm}
\usepackage{algorithm,algpseudocode}
\algnewcommand{\algorithmicforeach}{\textbf{for each}}
\algdef{SE}[FOR]{ForEach}{EndForEach}[1]
  {\algorithmicforeach\ #1\ \algorithmicdo}% \ForEach{#1}
  {\algorithmicend\ \algorithmicforeach}% \EndForEach
\usepackage{graphicx}
\usepackage{subfigure}
\newtheorem{theorem}{Theorem}
\newtheorem{lemma}{Lemma}

\title{SOAC: The Soft Option Actor-Critic Architecture}

\author{%
  Chenghao Li \\
  Tsinghua University \\
  lich18@mails.tsinghua.edu.cn\\
  \And
  Xiaoteng Ma \\
  Tsinghua University \\
  ma-xt17@mails.tsinghua.edu.cn \\
  \And
  Chongjie Zhang \\
  IIIS, Tsinghua University \\
  chongjie@tsinghua.edu.cn \\
  \AND
  Jun Yang\thanks{Corresponding author} \\
  Tsinghua University \\
  yangjun603@tsinghua.edu.cn \\
  \And
  Li Xia \\
  Sun Yat-Sen University \\
  xial@tsinghua.edu.cn \\
  \And
  Qianchuan Zhao \\
  Tsinghua University \\
  zhaoqc@tsinghua.edu.cn \\
}

\begin{document}

\maketitle

\begin{abstract}

The option framework has shown great promise by automatically extracting temporally-extended sub-tasks from a long-horizon task. Methods have been proposed for concurrently learning low-level intra-option policies and high-level option selection policy. However, existing methods typically suffer from two major challenges: ineffective exploration and unstable updates. In this paper, we present a novel and stable off-policy approach that builds on the maximum entropy model to address these challenges. Our approach introduces an information-theoretical intrinsic reward for encouraging the identification of diverse and effective options. Meanwhile, we utilize a probability inference model to simplify the optimization problem as fitting optimal trajectories. Experimental results demonstrate that our approach significantly outperforms prior on-policy and off-policy methods in a range of Mujoco benchmark tasks while still providing benefits for transfer learning. In these tasks, our approach learns a diverse set of options, each of whose state-action space has strong coherence.

%We also show that option selection policy can be transferred and accelerate learning in a new environment, even if the target task is dramatically different from the original one.

\end{abstract}

\section{Introduction}

%Such well-founded optimism depends largely on high-capacity function approximators such as neural networks. However, simply throwing in deep function approximators is far from adequate.

In the past few years, deep reinforcement learning (DRL) has shown remarkable progress in challenging application domains, such as Atari Games~\cite{mnih2015human}, Go game~\cite{silver2017mastering}, poker~\cite{brown2018superhuman}, StarCraft \uppercase\expandafter{\romannumeral2}~\cite{vinyals2019grandmaster}, and Dota 2~\cite{openai2019dota}. The combination of RL and high-capacity function approximators, such as neural networks, holds the promise of solving complex tasks in continuous control. However, millions of steps of data collection are needed to train effective behaviors. This training process might be simplified with a comprehensive understanding of tasks. A sophisticated agent should have the ability to identify distinct temporally-extended sub-tasks in a long-horizon task. How to efficiently discover such temporal abstractions has been widely studied in reinforcement learning (RL)~\cite{mcgovern2001automatic,barto2003recent,konidaris2009skill,da2012learning,kulkarni2016hierarchical,li2019hierarchical}. In this paper, we focus on the option framework~\cite{sutton1999between}, a distinct temporal abstraction method that can automatically discover courses of action with different intervals~\cite{riemer2018learning}. This distinct hierarchical structure has achieved notable success recently~\cite{bacon2017option,fox2017multi}.

%The majority of existing works focus on finding key states for the agent to reach~\cite{fishbach2006subgoals,vezhnevets2017feudal,levy2017learning,nachum2018data}. 

%Recently, a distinct
%%In this paper, we focus on 
%The option framework, a distinct hierarchical structure, 
%In this paper, we utilize a distinct view based on the option framework~\cite{sutton1999between}, which can automatically discover temporally abstract courses for learning~\cite{riemer2018learning}. Despite gradient-based option learning has achieved notable success recently~\cite{bacon2017option,fox2017multi}, there are still challenges hampering widespread adoption of the option framework.

However, there are remain challenges hampering widespread adoption of the option framework. One important such aspect is exploration. The option framework suffers from a degradation problem caused by ineffective exploration: there might be just one option selected to complete the entire task, which is tantamount to traditional end-to-end learning. Previous research tends to use on-policy learning to concurrently train option selection policy and intra-option policies~\cite{riemer2018learning,bacon2017option,fox2017multi,zhang2019dac}. However, only actually invoked options can be updated in on-policy learning. Intra-option policies sampled more frequently will be trained better to get more chance to be selected. This biased sampling makes the degradation problem worse. Another widely-known challenge is instability caused by simultaneous updates of high-level and low-level policies. Learning of intra-option policies will be unstable if the option selection policy frequently switches options to solve one sub-task. Previous work adapts option selection policy to updates of intra-option policies~\cite{zhang2019dac,osa2019hierarchical}. However, this short-sighted learning might exacerbate instability.

\begin{figure}[htbp]
\centering
\subfigure[Graphical model for a basic option trajectory.]{
\begin{minipage}[t]{0.5\linewidth}
\centering
\includegraphics[width=2.5in]{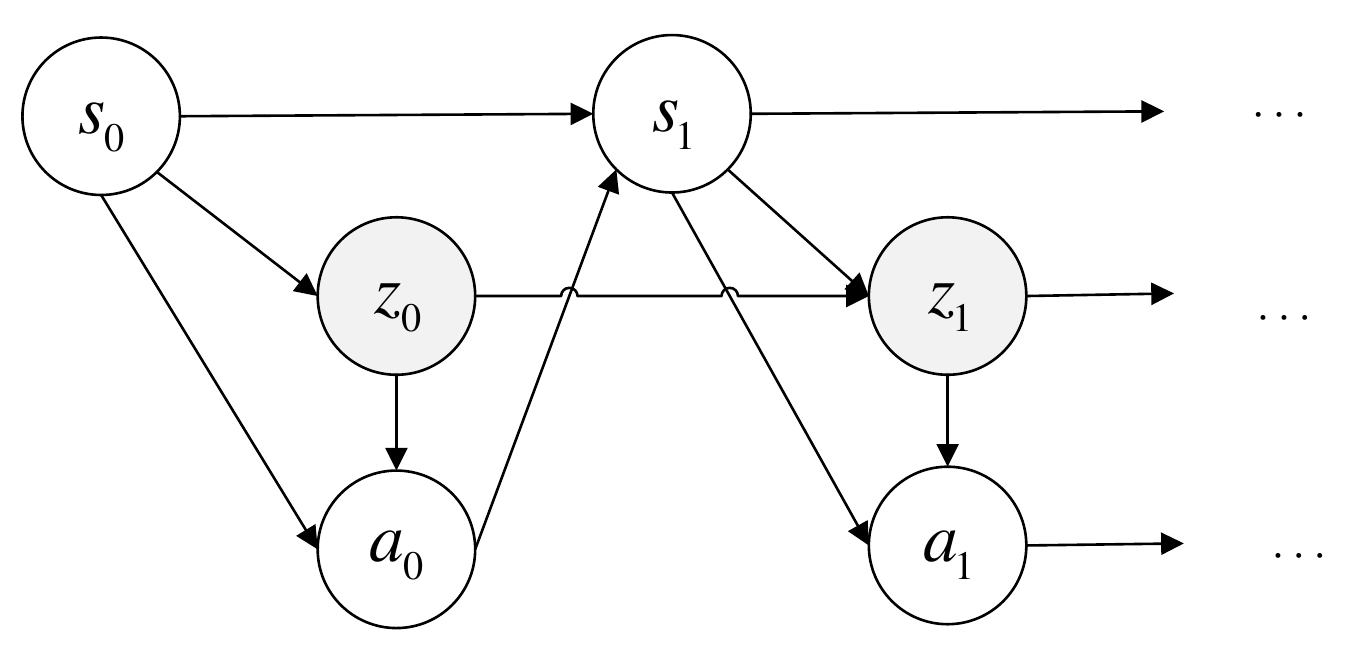}
%\caption{fig1}
\label{subfig:graph1}
\end{minipage}%
}%
\subfigure[Option trajectory with a probability inference model.]{
\begin{minipage}[t]{0.5\linewidth}
\centering
\includegraphics[width=2.5in]{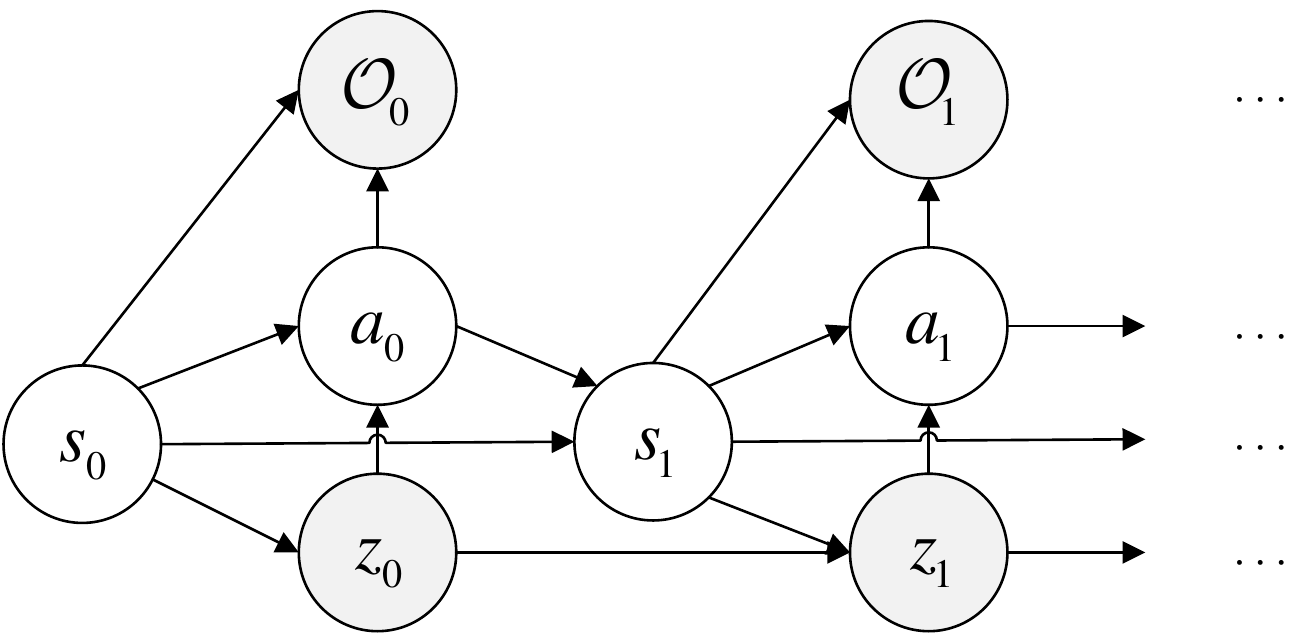}
\label{subfig:graph2}
%\caption{fig2}
\end{minipage}
}%
\centering
\caption{Grey nodes are hidden variables. {\bfseries Left}. The option framework introduces a hidden variable $z$ representing labels of intra-option policies. At each time step, option selection policy first decides whether to terminate the previous intra-option policy. If so, it will choose another intra-option policy depending on the current state. {\bfseries Right}. The optimality variable $\mathcal{O}$, theoretically indicating whether the current state-action pair is optimal, is introduced to the option framework. The hidden variable $z$ only affects the environment by guiding option selection policy to choose intra-option policies. So it not directly influences optimality variables. Instead, we utilize optimality variables to judge whether option selection policy is optimal, which is literately explained in Section~\ref{subsection:problem}.}
\label{fig:graph}
\end{figure}

To address these challenges, we present an off-policy soft option actor-critic (SOAC) approach that maximizes discounted rewards with entropy terms. This maximum entropy formulation provides sufficient exploration and robustness while acquiring diverse behaviors~\cite{haarnoja2017reinforcement,haarnoja2018soft}. The entropy bonus encourages the option selection policy to consider each intra-option policy in a balanced way. In addition, we introduce an information-theoretical intrinsic reward to enhance identifiability of intra-option policies. We utilize this intrinsic reward with another intrinsic reward related to anti-interference to define the objective for learning the optimal option selection policy. Meanwhile, we utilize external rewards to define the objective for learning the action selection policy. We theoretically derive that optimizing our maximum entropy model is equivalent to fitting optimal trajectories. Our algorithm can alternate between policy evaluation and policy improvement to learn optimal policies. Moreover, in our approach, the soft optimality of policies allows that behavior policies can be different from target policies~\cite{levine2018reinforcement,schulman2017equivalence}. With this flexibility, the option selection policy can be trained to select options with considering all historical behavior of each intra-option policy to reduce instability. As shown in Figure~\ref{fig:graph}, our algorithm learns a deep hierarchy of options.  

%Our approach called Soft Option Actor-Critic (SOAC) can seamlessly integrate SAC with the option framework to simplify the original task into sub-tasks for diverse options to learn. 

Experimental results indicate that our hierarchical approach significantly improves the performance of SAC~\cite{haarnoja2018soft} and outperforms state-of-the-art hierarchical RL algorithms~\cite{zhang2019dac,osa2019hierarchical} on the benchmark Mujoco tasks (Section~\ref{section:experiment}). In addition, we observe an obvious distinction between options, which indicates that a well trained option selection policy is sophisticated enough to invoke a diverse set of options in different situations. We also show that, the option selection policy can be transferred and accelerate learning in a new environment, even if the target task is dramatically different from the original one.

\section{Related Work}

Considerable prior work has explored how to extend the option framework~\cite{sutton1999between} to deep reinforcement learning (DRL). Compared with the end-to-end learning progress, learning the option framework from a single task brings more complex networks and more computational complexity. How to quickly learn an effective hierarchical structure is still an open question. Bacon et al.~\cite{bacon2017option} train the whole option framework with policy gradient method. To leverage recent advances in gradient-based policies, the option framework has combined with PPO~\cite{zhang2019dac,Schulman2017Proximal}, TD3~\cite{osa2019hierarchical,fujimoto2018addressing} and multitask~\cite{igl2019multitask}. To improve sample efficiency, all available intra-option policies can be trained simultaneously with a marginal distribution evaluating the probability of all options being selected~\cite{smith2018inference}. In addition, important sampling (IS) has been used to propose off-policy algorithms~\cite{harutyunyan2018learning,guo2017using} to reuse past experience. These research does show some interesting ways forward. However, they are not efficient enough compared with current baseline model-free DRL algorithms such as SAC~\cite{haarnoja2018soft}.

%Based on these research, we combine the option framework with probability inference models and an information-theoretical intrinsic reward to solve ineffective exploration and fragile updates.

%Meanwhile, to leverage recent advances in gradient-based policy, Zhang and Whiteson regard the whole option framework as two MDP problems and train them separately~\cite{zhang2019dac}. Osa et al. utilize mutual information to represent the identification of options with advantege-weighted importance~\cite{osa2019hierarchical}. Igl et al. regard the KL divergence between each task's policies and prior policies as a regular term to transfer information among multitask~\cite{igl2019multitask}.

Probability inference models provide a way to analyze the probability of optimal trajectories~\cite{levine2018reinforcement,kappen2012optimal,schulman2017equivalence}. Recently, these models have been adapted to numerous environments with DRL. Eysenbach et al.~\cite{eysenbach2018diversity} utilize diversity only to update parameters. Haarnoja et al.~\cite{haarnoja2018soft} propose the Soft Actor-Critic algorithm which is a state-of-the-art algorithm in single agent DRL. Huang et al.~\cite{huang2019svqn} optimize Partially Observable MDPs (POMDPs) with sequential variational soft Q-learning. Previous work based on soft optimality has shown both sample-efficient learning and stability. Meanwhile, the information bottleneck, related to mutual information (MI) and the Kullback-Leibler (KL) divergence, is widely used to control the spread of information~\cite{alemi2016deep,galashov2019information,goyal2019infobot,wang2019learning}. It can be used to judge division of state-action space~\cite{osa2019hierarchical} or used to distinguish different skills~\cite{sharma2019dynamics}. We propose our approach on these basis.

%propose a novel and stable off-policy algorithm to optimize the option framework with enough exploration.

%In this paper, we utilize the mutual information between options and state-action pair to reflect whether the option choosing policy is optimal.

%Our approach is closely related to Multitask Soft Option Learning~\cite{igl2019multitask} and Soft Options Critic~\cite{lobo2019soft}, which combine the option framework and the soft improvement model. The Multitask Soft Option Learning algorithm  Soft Options Critic algorithm maximize returns while maintaining a minimum target entropy. However, Multitask Soft Option Learning mainly focuses on how to transfer information among multitask. Our approach mainly focuses on how to learn the optimal option selection policy in the current task. In addition, experimental results in transfer environments indicate that our well trained option selection policy can accelerate training in a diametrically different task whether with or without previous trained intra-option policies. Meanwhile, the Soft Options Critic algorithm can be regarded as a special case of our algorithm while ignoring all the intrinsic reward related to the option selection policy. In addition, the lack of analysis of the optimal option selection policy will bring flaws to the theoretical derivation of probability inference models.

\section{Background}

\subsection{The Option Framework}

Traditional Markov Decision Process (MDP) considers a tuple $M=(\mathcal{S},\mathcal{A},P,R,\gamma)$. $\mathcal{S}$ is the state space, $\mathcal{A}$ is the action space, $P$ is the transition probability, $R$ is the relevant reward function, and $\gamma$ is a discount factor. The option framework extends the original MDP problem to a SMDP problem. It consists three components: a policy choosing options $\pi^o$, a termination condition $\beta$, and an initiation set $\mathcal{I}$~\cite{sutton1999between}. In this paper, we use $\mathcal{Z}$ to denote the option space. At each time step $t$, agents will decide whether to terminate the previous intra-option policy labeled as $z_{t-1}$ with the termination probability $\beta_{z_{t-1}}(s_t)$. If the previous intra-option policy is terminated, another intra-option policy will be sampled from $\pi^o\left(z_{t} | s_{t}\right)$. The whole probability of transitioning options written as below is called high-level option selection policy in this paper. Meanwhile each action is sampled from $\pi_{z_t}(\cdot|s_t)$ corresponding to the current option and state.

\begin{equation}
\tilde{\pi}_{\mathcal{Z}}\left(z_{t} | s_{t},z_{t-1}\right)=\left[1-\beta_{z_{t-1}}\left(s_{t}\right)\right] I_{z_{t}=z_{t-1}}+\beta_{z_{t-1}}\left(s_{t}\right) \pi^o\left(z_{t} | s_{t}\right).
\end{equation}

\subsection{Probability Inference Models}

Different from the general form of reinforcement learning problems, DRL based on probability inference models attempts to directly optimize the probability of optimal trajectories. An additional variable $\mathcal{O}_{t}$ is introduced to denote whether the current time step $t$ is optimal. This variable provides a mathematical formalization to analyze whether current policies are optimal. The log form of the probability of the optimal trajectories can be theoretically proved having an evidence low bound related to dense rewards and entropy~\cite{levine2018reinforcement}.

\begin{equation}
\begin{aligned}
\log p\left(\mathcal{O}_{1: T}\right) & \geq E_{\left({s}_{1: T}, {a}_{1: T}\right) \sim \pi \left({s}_{1: T}, {a}_{1: T}\right)}\left[\sum_{t=0}^{T} \left(r\left({s}_{t}, {a}_{t}\right)-\log \pi\left({a}_{t} | {s}_{t}\right)\right)\right]
\\ & = \sum_{t=0}^{T} E_{\left({s}_{t}, {a}_{t}\right) \sim \pi \left({s}_{t}, {a}_{t}\right)}\left[r\left({s}_{t}, {a}_{t}\right)+\mathcal{H}\left(\pi \left(\cdot | {s}_{t}\right)\right)\right],
\end{aligned}
\end{equation}

where $\pi (\cdot)$ is the actor policy, and $\mathcal{H}(\cdot)$ is the entropy regular term.

\section{Method}
\label{section:method}

In this section, we propose a maximum entropy problem and simplify it as fitting optimal trajectories with probability inference models. We propose an algorithm to estimate optimal policies iteratively.

%In addition, we define three value functions in log space to derive optimal policies and propose an algorithm to estimate them iteratively.

\subsection{Problem Formulation}
\label{subsection:problem}

Although previous research based on the option framework usually considers directly maximizing the reward function. We are interested in optimizing a maximum entropy model to solve the ineffective exploration challenge. In addition, we introduce mutual information $I\left(z_{t} ;\left\{s_{t}, a_{t}\right\} \right)$ as an intrinsic reward to enhance identifiability of each intra-option policy. Meanwhile, disturbance in a state-action pair should not lead to a substantial change in option selection~\cite{li2019hierarchical,puri2019explain}. So we add another intrinsic reward based on TV distance $\ell(\boldsymbol{\theta})={D}_{\mathrm{TV}}\left(p\left(z_{t} | s_{t}^{\text {noise }}, a_t^{\text {noise }} ; \boldsymbol{\theta}\right) \| p\left(z_{t} | s_{t}, a_{t} ; \boldsymbol{\theta}\right)\right)$ to encourage option selection policy to consider connectivity in state-action space while allocating options. In $\ell(\boldsymbol{\theta})$, $s_t^{\text {noise}}=s_t+\boldsymbol{\epsilon}_{\boldsymbol{s}}, a_t^{\text {noise }}=a_t+\boldsymbol{\epsilon}_{\boldsymbol{a}}$, $\boldsymbol{\epsilon}_{\boldsymbol{s}}$ and $\boldsymbol{\epsilon}_{\boldsymbol{a}}$ are gaussian noise, and $\boldsymbol{\theta}$ represents parameters of our model which can be neural networks. The whole maximum entropy problem is:

\begin{equation}
\begin{aligned} \pi^{H *}, \pi^{L *}=\underset{\pi^{H}, \pi^{L}}{\arg \max } \sum_{t} E_{\left(s_{t}, z_{t}, a_{t}\right) \sim \pi^{H}, \pi^{L}}&\left[\frac{r\left(s_{t}, a_{t}\right)}{\alpha}+\lambda_{1} I\left(z_{t} ;\left\{s_{t}, a_{t}\right\} \right)-\lambda_{2} \ell(\boldsymbol{\theta})\right.
\\ &\left.+\mathcal{H}\left(\pi^{H}\left(\cdot | s_{t}, z_{t-1}\right)\right)+\mathcal{H}\left(\pi^{L}\left(\cdot | s_{t}, z_{t}\right)\right)\right], \end{aligned}
\label{func:optimization}
\end{equation}

where we label high-level option selection policy as $\pi^H(z_t|s_t,z_{t-1})$ and label low-level intra-option policys as $\pi^L(a_t|s_t,z_t)$, $\mathcal{H}(\cdot)$ is entropy, $\alpha$ is a hyperparameter representing importance of external rewards, $\lambda_1$ and $\lambda_2$ are weights of intrinsic rewards.

To simplify the above problem, we introduce probability inference models. An additional variable $\mathcal{O}$ are introduced to describe whether the current condition is optimal. $\mathcal{O}_{t}=1$ indicates time step $t$ is optimal, and $\mathcal{O}_{t}=0$ indicates time step $t$ is not optimal. In the rest of this paper, we use $\mathcal{O}_{t}$ to represent $\mathcal{O}_{t}=1$ for concise functions. With this additional variable, we define a conditional probability model representing the probability of a trajectory $\tau$ with optimal policies:

\begin{equation}
\begin{aligned}
&p(\tau|\mathcal{O}_{0: T}) \propto p(\tau, \mathcal{O}_{0: T}) \propto p\left({s}_{0}\right) \prod_{t=0}^{T} p\left(\mathcal{O}_{t} | s_{t}, a_{t}\right) p(z_t|s_t,a_t,z_{t-1},\mathcal{O}_t) p\left({s}_{t+1} | {s}_{t}, {a}_{t}\right),
\label{optimal:prior}
\end{aligned}
\end{equation}

where $\mathcal{O}_{0: T}$ means $\mathcal{O}_{t}=1$ for all steps from $0$ to $T$. The probability of whether a state-action pair is optimal is defined as below, which is based on boltzmann distribution of energy~\cite{levine2018reinforcement}.

\begin{equation}
p\left(\mathcal{O}_{t} | s_{t}, a_{t}\right)=\exp \left(\frac{r\left(s_{t}, a_{t}\right)}{\alpha}\right).
\label{condition:0}
\end{equation}

Inspired by Equation~\ref{condition:0}, we utilize a similar exponential form to define the optimal option selection.

\begin{equation}
p(z_t|s_t,a_t,z_{t-1},\mathcal{O}_t)=\exp \left(\lambda_1 I(z_t;\{s_t,a_t\})-\lambda_2\ell(\boldsymbol{\theta})\right).
\label{condition:1}
\end{equation}

With option selection policy $\pi^H\left( \cdot | s_{t}\right)$ selecting options and intra-option policy $\pi^L\left(\cdot | {s}_{t},z_t\right)$ selecting actions, the probability of sampling a trajectory $\tau$ is:

\begin{equation}
{\hat{p}}({\tau})=p\left(s_{0}\right) \prod_{t=0}^{T} \pi^H\left(z_{t} | s_{t}, z_{t-1}\right) \pi^L \left(a_{t} | s_{t}, z_{t}\right) p\left(s_{t+1} | s_{t}, a_{t}\right).
\end{equation} 

\begin{theorem}

The original maximum entropy optimization problem shown in Equation\ref{func:optimization} can be simplified as shrinking the Kullback-Leibler (KL) divergence between $p\left(\tau | \mathcal{O}_{0: T}\right)$ and ${\hat{p}}({\tau})$.

\begin{equation}
\pi^{H *}, \pi^{L *}=\underset{\pi^{H}, \pi^{L}}{\arg \max }-{D}_{\mathrm{KL}}(\hat{p}(\tau) \| p(\tau| \mathcal{O}_{0: T})).
\end{equation}

\emph{Proof.}  See supplementary materials.

\end{theorem}

\subsection{Optimal Policies with Probability Inference Models}

In this sub-section, we derive optimal policies with probability inference models. First, we introduce three backward messages: $\beta_{t}\left(s_{t}\right)=p\left(\mathcal{O}_{t: T} | s_{t}\right)$, $\beta_{t}\left(s_{t}, z_{t}\right)=p\left(\mathcal{O}_{t: T} | s_{t}, z_{t}\right)$ and $\beta_{t}\left(s_{t}, z_{t}, a_{t}\right)=p\left(\mathcal{O}_{t: T} | s_{t}, z_{t}, a_{t}\right)$. These messages denote the probability of whether a trajectory starting from corresponding condition is optimal. With these backward messages, we can derive optimal option selection probability and optimal action selection probability as below.

\begin{equation}
p\left(z_{t} | s_{t}, \mathcal{O}_{t: T}\right) =\frac{p\left(s_{t}, z_{t} | \mathcal{O}_{t:T}\right)}{p\left(s_{t} | \mathcal{O}_{t: T}\right)}
=\frac{p\left(\mathcal{O}_{t: T} | s_{t}, z_{t}\right) p\left(z_{t} | s_{t}\right) p\left(s_{t}\right)}{p\left(\mathcal{O}_{t: T} | s_{t}\right) p\left(s_{t}\right)}  \propto \frac{p\left(\mathcal{O}_{t:T} | s_{t}, z_{t}\right)}{p\left(\mathcal{O}_{t: T} | s_{t}\right)}=\frac{\beta_{t}\left(s_{t}, z_{t}\right)}{\beta_{t}\left(s_{t}\right)},
\end{equation}

\begin{equation}
\begin{aligned}
p\left(a_{t} | s_{t}, z_{t}, \mathcal{O}_{t: T}\right)&=\frac{p\left(s_{t}, z_{t}, a_{t} | \mathcal{O}_{t:T}\right)}{p\left(s_{t}, z_{t} | \mathcal{O}_{t:T}\right)} =\frac{p\left(\mathcal{O}_{t:T} | s_{t}, z_{t}, a_{t}\right) p\left(a_{t} | s_{t}, z_{t}\right) p\left(s_{t},z_t\right)}{p\left(\mathcal{O}_{t:T} | s_{t},z_t\right) p\left(s_{t}, z_{t}\right)} \\& \propto \frac{p\left(\mathcal{O}_{t:T} | s_{t}, z_{t},a_t\right)}{p\left(\mathcal{O}_{t:T} | s_{t}, z_{t}\right)}=\frac{\beta_{t}\left(s_{t}, z_{t}, a_{t}\right)}{\beta_{t}\left(s_{t}, z_{t}\right)}.
\end{aligned}
\end{equation}

Inspired by Levine~\cite{levine2018reinforcement}, we use the log form of three backward messages to define value functions. We define $V\left(s_{t}\right)=\alpha \log \left(\beta_{t}\left(s_{t}\right) \right)$, $U\left(s_{t}, z_{t}\right)=\alpha \log \left(\beta_{t}\left(s_{t}, z_{t}\right) \right)$, and $Q\left(s_{t}, z_{t}, a_{t}\right)=\alpha \log \left(\beta_{t}\left(s_{t}, z_{t}, a_{t}\right) \right)$. With these value functions, optimal high level policy $\pi^{H *}$ and optimal low level policy $\pi^{L*}$ are derived as below.

\begin{equation}
\pi^{H *}=p\left(z_{t} | s_{t}, \mathcal{O}_{t: T}\right) \propto\frac{\beta_{t}\left(s_{t}, z_{t}\right)}{\beta_{t}\left(s_{t}\right)}=\frac{ \exp(\frac{1}{\alpha} U\left(s_{t}, z_{t}\right))}{\exp(\frac{1}{\alpha} V\left(s_{t}\right))},
\label{optimal:1}
\end{equation}

\begin{equation}
\pi^{L*}=p\left(a_{t} | s_{t}, z_{t}, \mathcal{O}_{t: T}\right)\propto\frac{\beta_{t}\left(s_{t}, z_{t}, a_{t}\right)}{\beta_{t}\left(s_{t}, z_{t}\right)}=\frac{ \exp(\frac{1}{\alpha} Q\left(s_{t}, z_{t}, a_t\right))}{\exp(\frac{1}{\alpha} U\left(s_{t}, z_{t}\right))},
\label{optimal:2}
\end{equation}

where $\alpha$ controls exploration degree. If $\alpha$ approaches infinity, optimal policies will obey uniform distribution. In contrast, if $\alpha$ approaches zero, optimal policies will be greedy. To estimate $V\left(s_{t}\right)$, $U\left(s_{t}, {z}_{t}\right)$ and $Q\left(s_{t}, {z}_{t}, a_{t}\right)$, we derive relationships between them.

\begin{lemma} 
\label{lemma1}
The relationship between $V\left(s_{t}\right)$ and $U\left(s_{t}, z_{t}\right)$ is:
\begin{equation}
\begin{aligned} 
V\left(s_{t}\right)=E_{\pi^{H*}(z_t | s_t)}\left[U\left(s_{t}, z_{t}\right)-\alpha \log \pi^{H*}\left(z_{t} | s_{t}\right)\right].
\end{aligned}
\label{lemma:1}
\end{equation}
\end{lemma}
\emph{Proof.}  See supplementary materials.

\begin{lemma} 
\label{lemma2}
The relationship between $U\left(s_{t}, z_{t}\right)$ and $Q\left(s_{t}, z_{t}, a_{t}\right)$ is:
\begin{equation}
\begin{aligned} 
U\left(s_{t}, {z}_{t}\right)=E_{\pi^{L*}\left(a_t | s_t, z_{t}\right)}\left[Q\left(s_{t}, z_{t}, a_{t}\right)-\alpha \log \pi^{L*}\left(a_{t} | s_{t}, z_{t}\right)\right].
\end{aligned}
\label{lemma:2}
\end{equation}
\end{lemma}
\emph{Proof.}  See supplementary materials.

\begin{lemma} 
\label{lemma3}
The relationship between $Q\left(s_{t}, z_{t}, a_{t}\right)$ and $V\left(s_{t+1}\right)$ is:

\begin{equation}
\begin{aligned} 
Q\left(s_{t}, {z}_{t}, a_{t}\right)=& r\left(s_{t}, a_{t}\right)+\alpha\left(\lambda_1 I(z_t;\{s_t,a_t\})-\lambda_2\ell(\boldsymbol{\theta})-\log p(z_t|s_t,a_t)\right)
\\ & +\gamma E_{p\left(s_{t+1} | s_{t}, a_{t}\right)}\left[V\left(s_{t+1}\right)\right].
\end{aligned} 
\label{lemma:3}
\end{equation}
\end{lemma}
\emph{Proof.}  See supplementary materials.

With these relationships between value functions, we can iteratively train them to estimate optimal policies. In the next sub-section, we will explain our algorithm in detail.

\subsection{Algorithm}

In this subsection, we will literally show our training process. We use function approximators and stochastic gradient descent to estimate and train U-value functions $U_{\phi_1}\left(s_t, z_t\right)$ and $U_{\phi_2}\left(s_t, z_t\right)$, Q-value functions $Q_{\theta_1}\left(s_t, z_t, a_t\right)$ and $Q_{\theta_2}\left(s_t, z_t, a_t\right)$, option selection policy $\pi_{\psi}^H(z_t|s_t,z_{t-1})$, and intra-option policys $\pi_{\zeta}^L(a_t|s_t,z_t)$. For more stable training, we utilize double neural networks~\cite{fujimoto2018addressing,van2016deep} and target neural networks~\cite{van2016deep,mnih2015human} while estimating U-value functions and Q-value functions. Q-value functions are trained by minimizing the Bellman residual shown as below, where we use the relationship between $V\left(s_{t}\right)$ and $U\left(s_{t}, z_{t}\right)$ shown in Equation~\ref{lemma:1}to replace $V\left(s_{t}\right)$.

\begin{equation}
  \label{alg:f1}
  \begin{aligned} 
  &J_{Q}(\theta_i) =E_{\left(s_{t}, z_{t}, a_{t},s_{t+1}\right) \sim \mathcal{D}}\left[\frac{1}{2}\left(Q_{\theta_i}\left(s_{t}, z_{t}, a_{t}\right)-\left(r\left(s_{t}, a_{t}\right)+\alpha\left(\lambda_1 I(z_t;\{s_t,a_t\})-\lambda_2\ell(\boldsymbol{\theta})\right.\right.\right.\right.
  \\ &\left.\left.\left.\left.-\log p(z_t|s_t,a_t)\right)+\gamma E_{z_{t+1} \sim \pi_{\psi}^H(.|s_{t+1})}\left[\min _{j=1,2}U_{\phi_j}\left(s_{t+1}, z_{t+1}\right)-\alpha \log \pi_{\psi}^H \left(z_{t+1} | s_{t+1},z_t\right)\right]\right)\right)^{2}\right].
  \end{aligned}
\end{equation}

The Bellman residual of U-value functions are:

\begin{equation}
\label{alg:f2}
\begin{aligned} J_{U}\left(\phi_{i}\right)=E_{\left(s_{t}, z_{t}\right) \sim \mathcal{D}}\left[\frac{1}{2}\left(U_{\phi_{i}}\left(s_{t}, z_{t}\right)-E_{a_{t} \sim \pi_{\zeta}^{L}\left(\cdot | s_{t}, z_{t}\right)}\right.\right.&\left[\min _{j=1,2} Q_{\theta_{j}}\left(s_{t}, z_{t}, a_{t}\right)\right.
\\ &\left.\left.\left.-\alpha \log \pi^{L}_{\zeta}\left(a_{t} | s_{t}, z_{t}\right)\right]\right)^{2}\right]. \end{aligned}
\end{equation}

It is difficult to directly calculate optimal high level policy $\pi_{\mathrm{new}}^{H *}$ and optimal low level policy $\pi_{\mathrm{new}}^{L*}$ from Equation~\ref{optimal:1} and Equation~\ref{optimal:2}. We use KL divergence to estimate policies. Option selection policy can be optimized by minimizing ${D}_{\mathrm{KL}}\left(\pi^{H}\left(\cdot | s_{t}, z_{t-1}\right) \| \pi^{H*}\right)$. Our option space $\mathcal{Z}$ is discrete. So we calculcate the expectation directly~\cite{christodoulou2019soft}.

\begin{equation}
\label{alg:f3}
\begin{aligned} 
J_{\pi^H}(\psi)&=E_{(s_{t},z_{t-1}) \sim D}\left[\pi_{\psi}^H\left(\cdot|s_t,z_{t-1}\right)^{T}\left(\alpha \log \left(\pi_{\psi}^H\left(.|s_t,z_{t-1}\right)\right)-\min _{i=1,2}U_{\phi_i}\left(s_{t},\cdot\right)\right)\right],
\end{aligned}
\end{equation}

where $\pi_{\psi}^H\left(\cdot|s_t,z_{t-1}\right)$ and $U_{\phi_i}\left(s_{t},\cdot\right)$ are the list of $\pi_{\psi}^H\left(z_t|s_t,z_{t-1}\right)$ and $U_{\phi_i}\left(s_{t},a_t\right)$, $\pi^H_{\psi}(z_t|s_t,z_{t-1})=\left(1-\beta_{\psi_1}\left(s_{t},z_{t-1}\right)\right) I_{z_{t}=z_{t-1}}+\beta_{\psi_1}\left(s_{t},z_{t-1}\right) \pi^o_{\psi_2}\left(z_{t} | s_{t}\right)$, $\beta_{\psi_1}\left(s_{t},z_{t-1}\right)$ decides whether to terminate previous options, $\pi^o_{\psi_2}\left(z_{t} | s_{t}\right)$ chooses new options, and $\psi = \{\psi_1,\psi_2\}$. Here both $\beta_{\psi_1}\left(s_{t},z_{t-1}\right)$ and $\pi^o_{\psi_2}\left(z_{t} | s_{t}\right)$ are trained by minimizing $J_{\pi^H}(\psi)$.

Intra-option policy is also optimized by minimizing $D_{\mathrm{KL}}\left(\pi^{L}\left(\cdot | s_{t}, z_{t}\right) \| \pi^{L*}\right)$. We use the reparameterization trick to allow gradients to pass through the expectations operator. At each time step $t$, ${a}_{t}$ is sampled from $f_{\zeta}\left(\epsilon_{t} ; {s}_{t},z_t\right)$, where $\epsilon_{t}$ is a noise vector sampled from a Gaussian distribution.

\begin{equation}
\label{alg:f4}
\begin{aligned} J_{\pi^{L}}(\zeta) &=E_{\left(s_{t}, z_{t}\right) \sim D, \epsilon_{t} \sim \mathcal{N}}\left[\alpha \log \left(\pi_{\zeta}^{L}\left(f_{\zeta}\left(\epsilon_{t} ; {s}_{t},z_t\right) | s_{t}, z_{t}\right)\right)-\min _{i=1,2}Q_{\theta_i}\left(s_{t}, z_{t}, f_{\zeta}\left(\epsilon_{t} ; {s}_{t},z_t\right)\right)\right]
\end{aligned}
\end{equation}

With all above loss functions, we can iteratively train value functions and estimate high-level and low-level optimal policies. The whole algorithm is literally listed in Algorithm~\ref{alg:example}.

\begin{algorithm}
  \caption{Soft Option Actor-Critic}
  \label{alg:example}
  \begin{algorithmic}[1]
    \State {\bfseries Input:} $\theta_1$, $\theta_2$, $\phi_1$, $\phi_2$, $\psi$, $\zeta$, $\alpha$ , $\tau$ \algorithmiccomment{Initialize parameters} 
    \\ $\bar{\theta}_{1} \leftarrow \theta_{1}$, $\bar{\theta}_{2} \leftarrow \theta_{2}$, $\bar{\phi}_{1} \leftarrow \phi_{1}$, $\bar{\phi}_{2} \leftarrow \phi_{2}$ \algorithmiccomment{Initialize target network weights}
    \\ $\mathcal{D} \leftarrow \emptyset$  \algorithmiccomment{Initialize an empty replay buffer}
    \For{each iteration}
    \For{each simulation step}
    \State $z_{t} \sim \pi^H_{\psi}\left(z_{t} | s_{t}, z_{t-1}\right)$, $a_{t} \sim \pi^L_{\zeta}\left(a_{t} | s_{t}, z_{t}\right)$, $s_{t+1} \sim p\left(s_{t+1} | s_{t}, a_{t}\right)$
    \State $\mathcal{D} \leftarrow \mathcal{D} \cup\left\{\left(z_{t-1},s_{t}, z_{t}, a_{t}, r_t, s_{t+1}\right)\right\}$
    \EndFor
    \For{each update step}
    \State $\phi_i \leftarrow \phi_i-\lambda_{U} \hat{\nabla}_{\phi_i} J_{U}\left(\phi_i\right)$, for $i \in\{1,2\}$
    \State $\theta_{i} \leftarrow \theta_{i}-\lambda_{Q} \hat{\nabla}_{\theta_{i}} J_{Q}\left(\theta_{i}\right)$, for $i \in\{1,2\}$
    \State $\psi \leftarrow \psi-\lambda_{\pi^H} \hat{\nabla}_{\psi} J_{\pi^H}\left(\psi\right)$, $\zeta \leftarrow \zeta-\lambda_{\pi^L} \hat{\nabla}_{\zeta} J_{\pi^L}\left(\zeta\right)$
    \EndFor
    \State $\bar{\theta}_{i} \leftarrow \tau\theta_{i}+(1-\tau)\bar{\theta}_{i}$, for $i \in\{1,2\}$ \algorithmiccomment{Soft update target network weights}
    \State $\bar{\phi}_{i} \leftarrow \tau\phi_{i}+(1-\tau)\bar{\phi}_{i}$, for $i \in\{1,2\}$\algorithmiccomment{Soft update target network weights}
    \EndFor
  \end{algorithmic}
\end{algorithm}

\section{Experiment}
\label{section:experiment}

In this section, we design experiments to answer following questions: (1) Can the additional option framework accelerate training? (2) Whether state-action space related to each option has strong coherence? (3) What is the impact of a well trained option selection policy in an opposite task? We adapt several benchmarking robot control tasks in Mujoco domains to answer the above questions.

\subsection{Results and Comparisons}
\label{exp:1}

We compare our algorithm with three other algorithms: Soft Actor-Critic (SAC)\cite{haarnoja2018soft}, Double Actor-Critic (DAC)~\cite{zhang2019dac} and adInfoHRL~\cite{osa2019hierarchical}. SAC is a current baseline off-policy DRL algorithm, which is also based on maximum entropy and probability inference models. We use it here to test whether our option framework can accelerate learning. Meanwhile, to the best of our knowledge, DAC and adInfoHRL are current best on-policy and off-policy algorithms with a similar hierarchical structure introducing a hidden and latent variable to abstractly present state-action space. All corresponding hyperparameters are literally listed in supplement materials.

\begin{figure}[H]\centering
\includegraphics[width=1\textwidth]{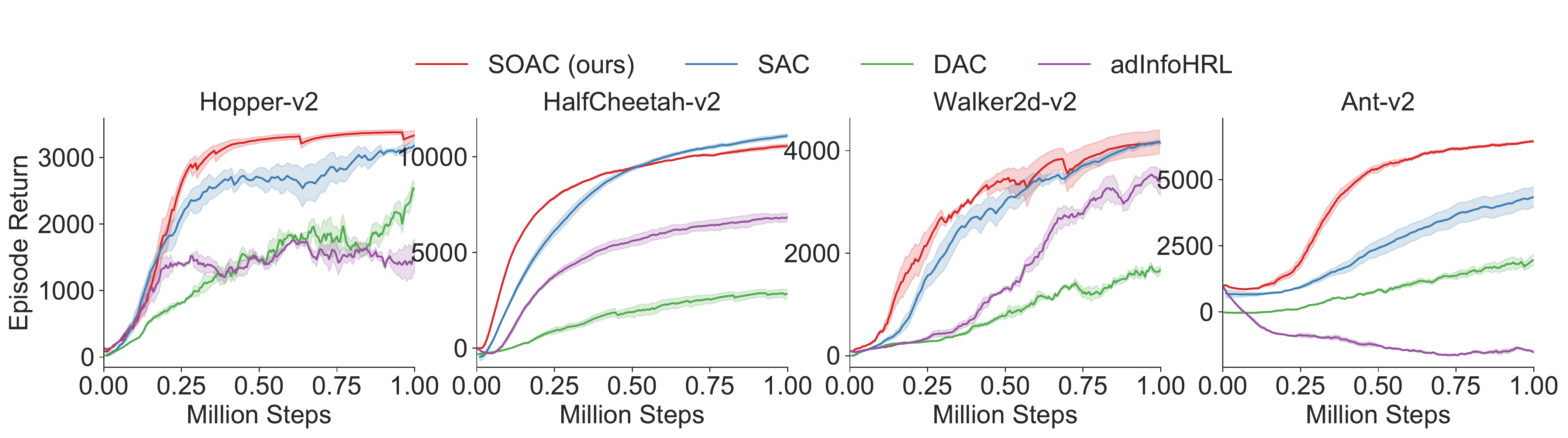}
\caption{Training curves of episode return in benchmark continuous control tasks.}
\label{fig:experiment1}
\end{figure}

Figure~\ref{fig:experiment1} demonstrates the average return of test rollout during training for SOAC (our algorithm), SAC, DAC and adInfoHRL on four Mujoco tasks. We train four different instances of each algorithm with random seeds from zero to three with each performing ten evaluation rollouts every 5000 environment steps and choose the best three instances. The solid curves represent the mean value smoothed by the Moving Average method and the shaded region represents the minimum and maximum returns over related trials. We notice that our algorithm dramatically outperforms DAC and adInfoHRL, both in terms of learning speed and stability. For example, on Hopper-v2, DAC and adInfoHRL suffer from unstable learning, but our algorithm quickly stabilizes at the highest score. Meanwhile, on Ant-v2, addInfoRL fails to make any progress, but our algorithm dramatically outperforms other algorithms. Compared with SAC, our algorithm performs comparably on HalfCheetah-v2 and Walker2d-v2 and outperforms on Hopper-v2 and Ant-v2. These results indicate that our algorithm can accelerate learning by softly dividing state-action space based on the option framework with sufficient exploration. We address part of the reason as the multimodal treatment of our actor's policy. To deal with continuous action space, an actor's policy is usually defined as a normal distribution. However, this might not meet the actual optimal policy. The entire policy of our actor has a multimodal distribution similar to Gaussian Mixture Model (GMM) and give our agents a stronger ability to make decisions. In addition, part of neural networks related to different intra-option policies are shared to accelerate training~\cite{zheng122018self}. This provides the same feature extraction strategy for all intra-option policies.

\subsection{Visualization of State-Action Space with Different Options}

\begin{figure}[H]\centering
\includegraphics[width=.245\textwidth]{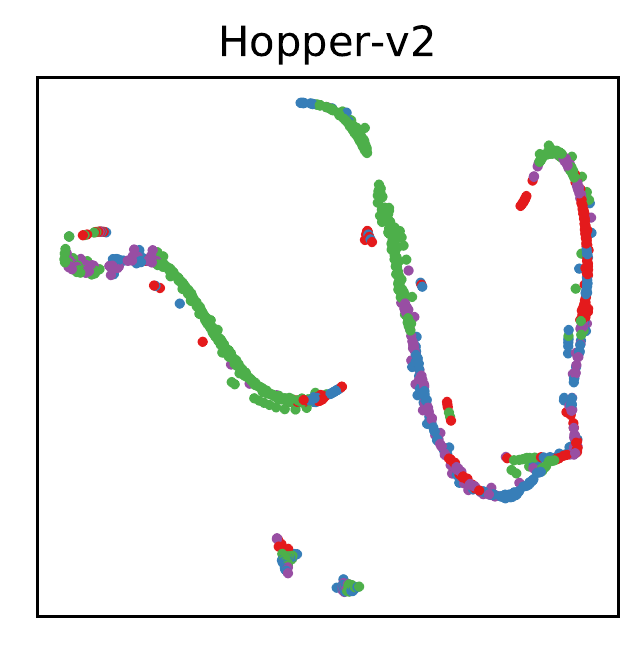}
\includegraphics[width=.245\textwidth]{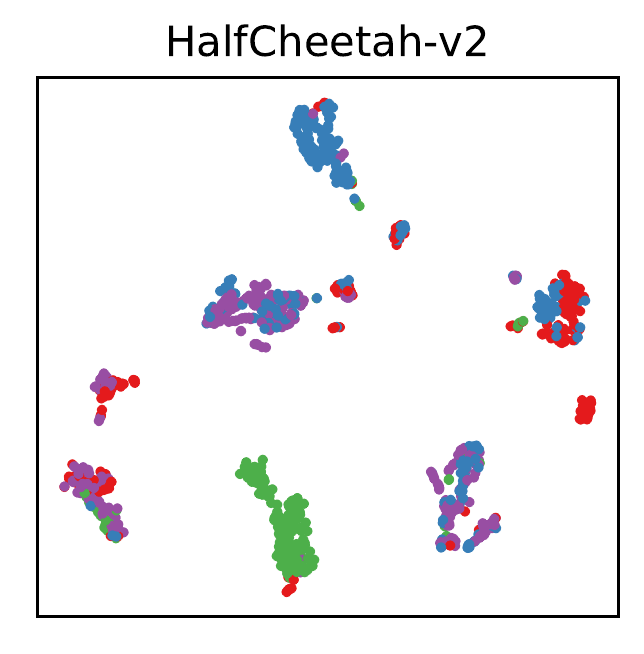}
\includegraphics[width=.245\textwidth]{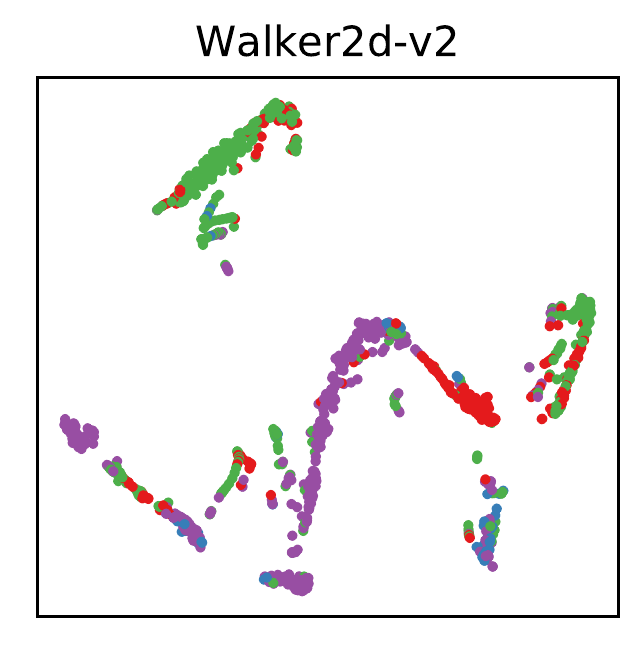}
\includegraphics[width=.245\textwidth]{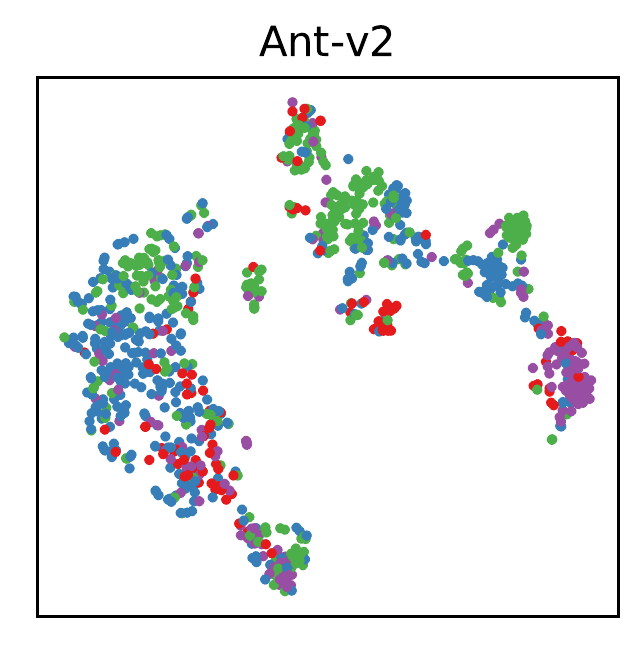}
\caption{Embeddings visualizing of state-action space with the t-sne method.}
\label{fig:tsne}
\end{figure}

Our algorithm performs well in Mujoco domains with stable learning curves. To verify whether our option selection policy is reasonable, we utilize the t-sne method~\cite{maaten2008visualizing} to illustrate state-action space corresponding to each option in Figure~\ref{fig:tsne}. We notice distinct clusters for different options in each Mujoco task. This indicates that our option selection policy is well trained to assign options for different situations. In addition, we notice multi-cluster related to one option, which is similiar with~\cite{osa2019hierarchical,oord2018representation,goyal2019reinforcement}. This is because option selection policy might assign different sub-tasks for one option to solve limited by the fixed number of options. How to determine the most suitable number of options is still an open question, although most previous research tends to set the number to four~\cite{bacon2017option,smith2018inference,osa2019hierarchical,zhang2019dac}. An exploration of variable number of options is a future direction.

\subsection{Transfer of Option Selction Policy}
\label{exp:2}

%Distinct from other hierarchical structures, option selection policy in the option framework is more like a problem maker rather than an organizer. 

\begin{figure}[H]\centering
\includegraphics[width=1\textwidth]{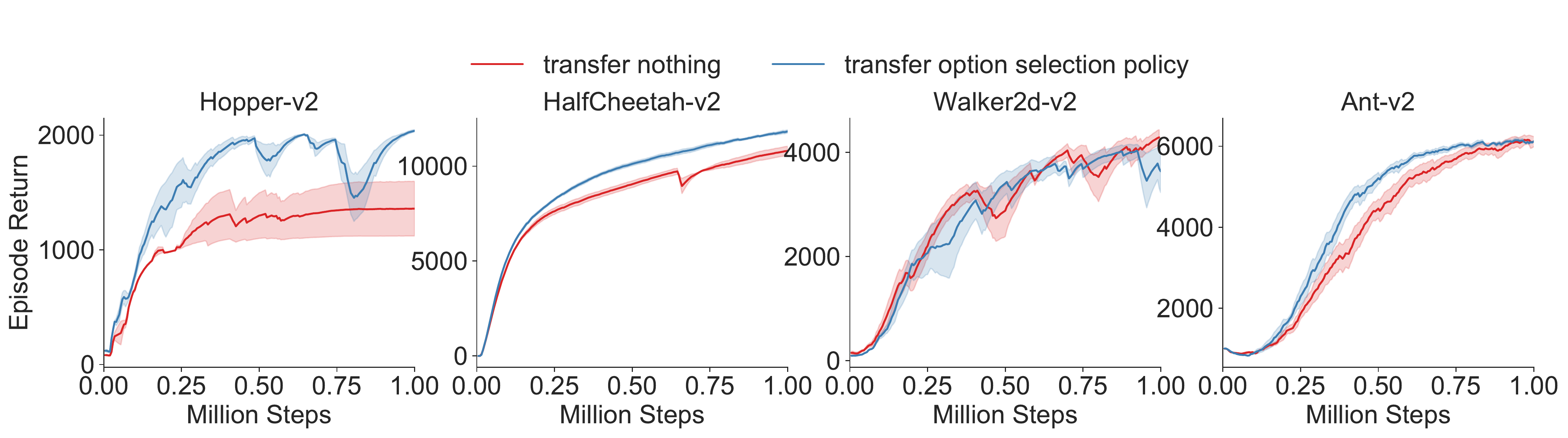}
\caption{Learning curves of transferring option selection policy compared with transferring nothing.}
\label{fig:transfer}
\end{figure}

We are wondering whether the option selection policy learned from a certain task can discover a general division method of the environment. Even though our algorithm is not designed for transfer learning, we find out that our well trained option selection policy can accelerate training in a diametrically different task with an \textbf{opposite} reward function. As shown in Figure~\ref{fig:transfer}, transferring high-level option selection policy will accelerate learning compared with transferring nothing in most Mujoco domains. Meanwhile, the transferred option selection policy makes the training more stable. Especially on Hopper-v2, a hopper suffers from falling down while attempting to jump backwards. With transferred option selection policy, agents have more opportunities to learn to jump backwards rather than staying in place. These results indicate that our well trained option selection policy can generally divide the environment and assign sub-tasks with probability models, which will provide benefits for transfer learning.

%Although for each transfer task, there is a poor start of episode return by transferring both $\pi^H$ and $\pi^L$, agents will quickly adjust themselves to adapt to a new task as shown in Figure~\ref{fig:transfer}.

\section{Conclusion}

In this paper, we propose soft option actor-critic (SOAC), an off-policy maximum entropy DRL algorithm with the option framework. With probability inference models, we theoretically derive optimal policies based on soft optimality and simplify our optimization problem as fitting optimal trajectories. We empirically demonstrate that our algorithm matches or exceeds prior on-policy and off-policy methods in a range of Mujoco benchmark tasks and while still providing benefits for transfer learning. The state-action space associated with each option shows strong connectivity. These results indicate that our option selection policy is sophisticated to assign options for different situation. Our algorithm has shown the potential to boost sample efficiency with operative exploration to address current well-known challenges restricting the applicability of the option framework. 

%\newpage

\section{Broader Impact}

Deep reinforcement learning (DRL) has achieved remarkable progress in recent years. It has exceeded the human level performance in many challenging environments such as Atari Games~\cite{mnih2015human,Mnih2013Playing}, game of go~\cite{silver2017mastering}, poker~\cite{brown2018superhuman}, and StarCraft \uppercase\expandafter{\romannumeral2}~\cite{vinyals2019grandmaster}. However, classical end-to-end learning progress still suffers from high dimension in state and action space, which might influence the convergence rate and cause unbearable training time. In this paper, we attempt to train an option framework, which can extract sub-tasks with arbitrary interval from a long-horizon task to simplify the original MDP problem. We combine the option framework with probability inference models and information-theoretical intrinsic rewards and propose a novel and stable off-policy algorithm to address the well known challenges mentioned in the introduction section. As we all know, creation starts from the ability to discover and summarize problems. With the option framework, agents can learn diverse skills from sub-tasks proposed by themselves while solving the entire task. In general, the option framework encourages agents to explore the environment and ask questions. This might be a key point in the artificialization of artificial intelligence. Learning the option framework will definitely bring more computational complexity. Nevertheless, our approach has shown that learning this hierarchical structure can accelerate training in Mujoco domains. Our approach can be regarded as a step for the option framework to be widespreadly adopted.

%\nocite{kingma2014adam,hessel2019multi}

%\bibliographystyle{unsrt}
\bibliographystyle{authordate1}
%\bibliography{references}{}
\bibliography{SOAC}{}
\bibliographystyle{plain}

\newpage

\begin{appendix}

\section{Theory Details}

\subsection{Graphical Models}

The whole trajectory is shown in Figure.~\ref{fig:graph}. Its corresponding distribution $p(\tau)$ is:

\begin{equation}
p(\tau) =p\left(s_{0}, z_{0}, a_{0}, \ldots, s_{T}, z_{T}, a_{T} | \theta\right) =p\left(s_{0}\right) \prod_{t=0}^{T} p\left(a_{t} | s_{t}, z_{t}, \theta\right) p\left(z_{t} | s_{t},z_{t-1}, \theta\right) p\left(s_{t+1} | s_{t}, a_{t}\right),
\end{equation}

where $p\left(z_{t} | s_{t},z_{t-1},\theta \right)=\left(1-\beta_{z_{t-1}}\left(s_{t}|\theta\right)\right) I_{z_{t}=z_{t-1}}+\beta_{z_{t-1}}\left(s_{t}|\theta\right) \pi^o\left(z_{t} | s_{t},\theta\right)$, $\beta_{z_{t-1}}\left(s_{t}|\theta\right)$ is a terminal condition function, and $\pi^o\left(z_{t} | s_{t},\theta\right)$ is an option choosing policy.

\subsection{Derivation of the Optimization Problem}

Based on probability models corresponding to $\mathcal{O}_{t}$ shown in Equation~\ref{condition:0} and Equation~\ref{condition:1}, we can recover the explicit form of $p\left(\tau | \mathcal{O}_{1: T}\right)$ from Equation~\ref{optimal:prior}.

\begin{equation}
\begin{aligned}
p(\tau|\mathcal{O}_{1:T}) & \propto p\left({s}_{0}\right) \prod_{t=0}^{T} p\left(\mathcal{O}_{t} | s_{t}, a_{t}\right) p(z_t|s_t,a_t,z_{t-1},\mathcal{O}_t) p\left({s}_{t+1} | {s}_{t}, {a}_{t}\right) 
\\ & = \mathcal{M} p\left({s}_{0}\right) \prod_{t=0}^{T} p\left(\mathcal{O}_{t} | s_{t}, a_{t}\right) p(z_t|s_t,a_t,z_{t-1},\mathcal{O}_t) p\left({s}_{t+1} | {s}_{t}, {a}_{t}\right) 
\\ & = \mathcal{M} \left[p\left({s}_{0}\right) \prod_{t=0}^{T} p\left({s}_{t+1} | {s}_{t}, {a}_{t}\right)\right] \exp \left(\sum_{t=0}^{T}\frac{r\left(s_{t}, a_{t}\right)}{\alpha}\right)
\\ & \qquad\qquad\qquad\qquad\qquad\qquad\qquad\exp \left(\sum_{t=0}^{T}\left(\lambda_1 I(z_t;\{s_t,a_t\})-\lambda_2\ell(\boldsymbol{\theta})\right)\right),
\label{optimal:prior_expend}
\end{aligned}
\end{equation}

where $\mathcal{M}$ is a constant representing the multiplication of some prior probabilities.

Our optimization process can be defined as continuously shrinking the KL divergence from the optimal strategy, which can be written as

\begin{equation}
\begin{aligned}
{D}_{\mathrm{KL}}({\hat{p}}({\tau}) \| p(\tau|\mathcal{O}_{1:T})) &=E_{\tau \sim \hat{p}(\tau)}\left[\log \hat{p}(\tau)-\log {p}(\tau|\mathcal{O}_{1:T})\right]
\\ &=\sum_{t=0}^{T}E_{(s_t,z_t,a_t) \sim \hat{p}(\tau)}\left[ \log \pi^H\left(z_{t} | s_{t}, z_{t-1}\right) + \log \pi^L \left(a_{t} | s_{t}, z_{t}\right) -\frac{r\left(s_{t}, a_{t}\right)}{\alpha}\right.
\\ &\left.\qquad\qquad\qquad\qquad\quad-\left(\lambda_{1} I\left(z_{t} ;\left\{s_{t}, a_{t}\right\} \right)-\lambda_{2} \ell(\boldsymbol{\theta})\right)\right]-\log \mathcal{M},
\end{aligned}
\end{equation}

where $\log \mathcal{M}$ is a constant which can be ignored while optimizing policies to maximize or minimize ${D}_{\mathrm{KL}}({\hat{p}}({\tau}) \| p(\tau|\mathcal{O}_{1:T}))$. Our optimization problem can be further simplified to:

\begin{equation}
\begin{aligned}
\pi^{H*}, \pi^{L*} &= \arg \max _{\pi^H, \pi^L} -{D}_{\mathrm{KL}}({\hat{p}}({\tau}) \| p(\tau| \mathcal{O}_{0: T})) 
\\ &= \arg \max _{\pi^H, \pi^L} \sum_{t=0}^{T}E_{(s_t,z_t,a_t)  \sim \hat{p}(\tau)}\left[ \frac{r\left(s_{t}, a_{t}\right)}{\alpha}+\left(\lambda_{1} I\left(z_{t} ;\left\{s_{t}, a_{t}\right\} \right)-\lambda_{2} \ell(\boldsymbol{\theta})\right)\right.
\\ &\left.\qquad\qquad\qquad\qquad\qquad\qquad\quad -\log \pi^H\left(z_{t} | s_{t}, z_{t-1}\right) - \log \pi^L \left(a_{t} | s_{t}, z_{t}\right) \right].
\\ &= \arg \max _{\pi^H, \pi^L} \sum_{t=0}^{T}E_{(s_t,z_t,a_t)  \sim \hat{p}(\tau)}\left[ \frac{r\left(s_{t}, a_{t}\right)}{\alpha}+\left(\lambda_{1} I\left(z_{t} ;\left\{s_{t}, a_{t}\right\} \right)-\lambda_{2} \ell(\boldsymbol{\theta})\right)\right.
\\ &\left.\qquad\qquad\qquad\qquad\qquad\qquad\quad +\mathcal{H}\left(\pi^{H}\left(\cdot | s_{t}\right)\right)+\mathcal{H}\left(\pi^{L}\left(\cdot | s_{t}, z_{t}\right)\right) \right].
\\ &= \arg \max _{\pi^H, \pi^L} \sum_{t=0}^{T}E_{(s_t,z_t,a_t) \sim \pi^{H}, \pi^{L}}\left[ \frac{r\left(s_{t}, a_{t}\right)}{\alpha}+\left(\lambda_{1} I\left(z_{t} ;\left\{s_{t}, a_{t}\right\} \right)-\lambda_{2} \ell(\boldsymbol{\theta})\right)\right.
\\ &\left.\qquad\qquad\qquad\quad\qquad\qquad\qquad\quad +\mathcal{H}\left(\pi^{H}\left(\cdot | s_{t}\right)\right)+\mathcal{H}\left(\pi^{L}\left(\cdot | s_{t}, z_{t}\right)\right) \right].
\end{aligned}
\end{equation}

\subsection{Relationship among Backward Messages}

The relationship among $\beta_{t}\left(s_{t}\right)$, $\beta_{t}\left(s_{t}, z_{t}\right)$ and $\beta_{t}\left(s_{t}, z_{t}, a_{t}\right)$ is:

\begin{equation}
\beta_{t}\left(s_{t}\right)=p\left(\mathcal{O}_{t: T} | s_{t}\right)=\int_{\mathcal{Z}} p\left(\mathcal{O}_{t: T} | s_{t}, z_{t}\right) p\left(z_{t} | s_{t}\right) d z_{t}=\int_{\mathcal{Z}} \beta_{t}\left(s_{t}, z_{t}\right) p\left(z_{t} | s_{t}\right) d z_{t},
\label{relation:1}
\end{equation}

where $p\left(z_{t} | s_{t}\right)$ is the prior option choosing policy and can be assumed as a uniform distribution over the set of option.

\begin{equation}
\begin{aligned}
\beta_{t}\left(s_{t}, z_{t}\right)=p\left(\mathcal{O}_{t: T} | s_{t}, z_{t}\right) & =\int_{\mathcal{A}} p\left(\mathcal{O}_{t: T} | s_{t}, z_{t}, a_{t}\right) p\left(a_{t} | s_{t}, z_{t}\right) d a_{t}
\\ & = \int_{\mathcal{A}} \beta_{t}\left(s_{t}, z_{t}, a_{t}\right) p\left(a_{t} | s_{t}, z_{t}\right) d a_{t},
\end{aligned}
\label{relation:2}
\end{equation}

where $p\left(a_{t} | s_{t}, z_{t}\right)$ is the prior action choosing policy and can be assumed as a uniform distribution over the set of action.

\begin{equation}
\begin{aligned}
\beta_{t}\left(s_{t}, z_{t}, a_{t}\right)=p\left(\mathcal{O}_{t: T} | s_{t}, z_{t}, a_{t}\right)&=\int_{\mathcal{S}} p\left(\mathcal{O}_{t+1: T} | s_{t+1}\right)  p\left(s_{t+1} | s_{t}, z_{t}, a_{t}\right) p\left(\mathcal{O}_{t} | s_{t}, z_{t}, a_{t}\right) d s_{t+1}
\\ & =\int_{\mathcal{S}} \beta_{t+1}\left(s_{t+1}\right) p\left(s_{t+1} | s_{t}, a_{t}\right) p\left(\mathcal{O}_{t} | s_{t}, z_{t}, a_{t}\right) d s_{t+1}.
\end{aligned}
\label{relation:3}
\end{equation}

\subsection{Proof of Lemma 1}
\label{sec:l1}

\begin{proof}\renewcommand{\qedsymbol}{}

We assume the prior option choosing policy is equally probable in all possible values. To simplify our formulation, we assume the value of $p(z_t|s_t)$ is one no matter what $z_t$ is. This might cause the estimated V function to be a multiple of the actual V function. Our optimal option choosing policy and optimal action choosing policy have the softmax form. So this multiple form error will not lead to changes in the optimal policies. In addition, we believe a sophisticated alpha can offset the deviation. Based on assumptions above, the V function can be written as:

\begin{equation}
\begin{aligned} 
V\left(s_{t}\right) &= \alpha \log \int_{\mathcal{Z}} \exp \left(\frac{U\left(s_{t}, z_{t}\right)}{\alpha}\right)p(z_t|s_t) d z_{t}
\\ &\overset{def}{=} \alpha \log \int_{\mathcal{Z}} \exp \left(\frac{U\left(s_{t}, z_{t}\right)}{\alpha}\right) d z_{t}
\\ &=E_{\pi^{H*}\left(z_t | s_{t}\right)}\left[U\left(s_{t}, z_{t}\right)-U\left(s_{t}, z_{t}\right)+\alpha \log \int_{\mathcal{Z}} \exp \left(\frac{U\left(s_{t}, z_{t}\right)}{\alpha}\right) d z_{t} \right] 
\\ &=E_{\pi^{H*}\left(z_t | s_{t}\right)}\left[U\left(s_{t}, z_{t}\right)-\alpha \log \exp \frac{U\left(s_{t}, z_{t}\right)}{\alpha}+\alpha \log \int_{\mathcal{Z}} \exp \left(\frac{U\left(s_{t}, z_{t}\right)}{\alpha}\right) d z_{t} \right] 
\\ &=E_{\pi^{H*}\left(z_t | s_{t}\right)}\left[U\left(s_{t}, z_{t}\right)-\alpha \log \frac{\exp \frac{U\left(s_{t}, z_{t}\right)}{\alpha}}{\int_{\mathcal{Z}} \exp \left(\frac{U\left(s_{t}, z_{t}\right)}{\alpha}\right) d z_{t}}\right] 
\\ &=E_{\pi^{H*}\left(z_t | s_{t}\right)}\left[U\left(s_{t}, z_{t}\right)-\alpha \log \frac{\exp \frac{U\left(s_{t}, z_{t}\right)}{\alpha}}{\exp \frac{V\left(s_{t}\right)}{\alpha}}\right] 
\\ &=E_{\pi^{H*}(z_t | s_t)}\left[U\left(s_{t}, z_{t}\right)-\alpha \log \pi^{H*}\left(z_{t} | s_{t}\right)\right].
\end{aligned}
\end{equation}

\end{proof}

\subsection{Proof of Lemma 2}

\begin{proof}\renewcommand{\qedsymbol}{}

Same as Section~\ref{sec:l1}, we set the value of $p(a_t|s_t,z_t)$ to one no matter what $a_t$ is. Then the U function can be written as:

\begin{equation}
\begin{aligned} 
U\left(s_{t}, z_{t}\right) &= \alpha \log \int_{A} \exp \left(\frac{Q\left(s_{t}, z_{t}, a_{t}\right)}{\alpha}\right) p(a_t|s_t,z_t) d a_{t} 
\\ &\overset{def}{=}\alpha \log \int_{A} \exp \left(\frac{Q\left(s_{t}, z_{t}, a_{t}\right)}{\alpha}\right) d a_{t} 
\\ &=E_{\pi^{L*}\left(a_t | s_t, z_{t}\right)}\left[Q\left(s_{t}, z_{t}, a_{t}\right)-Q\left(s_{t}, z_{t}, a_{t}\right)+\alpha \log \int_{A} \exp \left(\frac{Q\left(s_{t}, z_{t}, a_{t}\right)}{\alpha}\right) d a_{t}\right] 
\\ &=E_{\pi^{L*}\left(a_t | s_t, z_{t}\right)}\left[Q\left(s_{t}, z_{t}, a_{t}\right)-\alpha \log \exp \frac{Q\left(s_{t}, z_{t},a_t\right)}{\alpha}+\alpha \log \int_{A} \exp \left(\frac{Q\left(s_{t}, z_{t}, a_{t}\right)}{\alpha}\right) d a_{t}\right] 
\\ &=E_{\pi^{L*}\left(a_t | s_t, z_{t}\right)} \left[Q\left(s_{t}, z_{t}, a_{t}\right)-\alpha \log \frac{\exp \frac{Q\left(s_{t}, z_{t},a_t\right)}{\alpha}}{\int_{\mathcal{A}} \exp \left(\frac{Q\left(s_{t}, z_{t},a_t\right)}{\alpha}\right) d a_{t}}\right] 
\\ &=E_{\pi^{L*}\left(a_t | s_t, z_{t}\right)}\left[Q\left(s_{t}, z_{t}, a_{t}\right)-\alpha \log \frac{\exp \frac{Q\left(s_{t}, z_{t},a_t\right)}{\alpha}}{\exp \frac{U\left(s_{t},a_t\right)}{\alpha}}\right] 
\\ &=E_{\pi^{L*}\left(a_t | s_t, z_{t}\right)}\left[Q\left(s_{t}, z_{t}, a_{t}\right)-\alpha \log \pi^{L*}\left(a_{t} | s_{t}, z_{t}\right)\right].
\end{aligned}
\end{equation}

\end{proof}

\subsection{Proof of Lemma 3}

\begin{proof}\renewcommand{\qedsymbol}{}

Because of the relationship between $\beta_{t}\left(s_{t}, z_{t}, a_{t}\right)$ and $\beta_{t+1}\left(s_{t+1}\right)$ shown in Equation\ref{relation:3}, the Q function can be written as below. The original backup considers a softmax over the next expected value. In that way, one possible outcome for the next state with a very high value will dominate the backup~\cite{levine2018reinforcement}. So we replace the softmax form backup message $\alpha \log E_{p\left(s_{t+1} | s_t, a_{t}\right)}\left[\exp \left(\frac{V\left(s_{t+1}\right)}{\alpha}\right)\right]$ with a more general form: $\gamma E_{p\left(s_{t+1} | s_{t}, a_{t}\right)}\left[V\left(s_{t+1}\right)\right]$.

\begin{equation}
\begin{aligned} 
Q\left(s_{t}, z_{t}, a_{t}\right) &= \alpha \log \int_{\mathcal{S}} \exp \left(\frac{V\left(s_{t+1}\right)}{\alpha}\right) p\left(s_{t+1} | s_{t}, a_{t}\right) p\left(\mathcal{O}_{t} | s_{t}, z_{t}, a_{t}\right) d s_{t+1}
\\ & = \alpha \log E_{p\left(s_{t+1} | s_t, a_{t}\right)}\left[\exp \left(\frac{V\left(s_{t+1}\right)}{\alpha}\right)\right] + \alpha \log p\left(\mathcal{O}_{t} | s_{t}, z_{t}, a_{t}\right)
\\ & \overset{def}{=} \gamma E_{p\left(s_{t+1} | s_{t}, a_{t}\right)}\left[V\left(s_{t+1}\right)\right] + \alpha \log p\left(\mathcal{O}_{t} | s_{t}, z_{t}, a_{t}\right)
\\ & = \gamma E_{p\left(s_{t+1} | s_{t}, a_{t}\right)}\left[V\left(s_{t+1}\right)\right] + \alpha \log \frac{p\left(\mathcal{O}_t, s_{t},z_{t},a_{t}\right)}{p\left(s_{t},z_{t},a_t\right)}
\\ & = \gamma E_{p\left(s_{t+1} | s_{t}, a_{t}\right)}\left[V\left(s_{t+1}\right)\right] + \alpha \log \frac{p\left(z_t|s_t,a_t,\mathcal{O}_t\right) p\left(\mathcal{O}_t| s_{t},a_{t}\right) p\left(s_{t},a_{t}\right)}{p\left(z_{t}|s_{t},a_t\right)p\left(s_{t},a_t\right)}
\\ & = \gamma E_{p\left(s_{t+1} | s_{t}, a_{t}\right)}\left[V\left(s_{t+1}\right)\right] + \alpha \log \frac{p\left(z_t|s_t,a_t,\mathcal{O}_t\right) p\left(\mathcal{O}_t| s_{t},a_{t}\right)}{p\left(z_{t}|s_{t},a_t\right)}
\\ & = \gamma E_{p\left(s_{t+1} | s_{t}, a_{t}\right)}\left[V\left(s_{t+1}\right)\right] + \alpha \log  p\left(\mathcal{O}_t| s_{t},a_{t}\right)+ \alpha \log p\left(z_t|s_t,a_t,\mathcal{O}_t\right)- \alpha \log p\left(z_{t}|s_{t},a_t\right)
\\ & \overset{def}{=} \gamma E_{p\left(s_{t+1} | s_{t}, a_{t}\right)}\left[V\left(s_{t+1}\right)\right]+r\left(s_{t}, a_{t}\right)+\alpha\left(\lambda_1 I(z_t|s_t,a_t)-\lambda_2\ell(\boldsymbol{\theta})-\log p(z_t|s_t,a_t)\right)
\end{aligned}
\end{equation}

\end{proof}

\subsection{Mutual Information Calculation}

In this paper, we utilize the mutual information (MI) $I\left(z_{t} ;\left\{s_{t}, a_{t}\right\}\right)$ to describe the identifiability of each intra-option policy, which is defined as

\begin{equation}
I\left(z_{t} ;\left\{s_{t}, a_{t}\right\}\right)=H(z_t)-H(z_t | s_t, a_t).
\end{equation}

To calculate entropy, we need to calcutate corresponding probability $p(z_t | s_t, a_t) $ and $p(z_t)$ first. $p(z_t | s_t, a_t)$ is calculated by Bayesian formula. And $p(z_t)$ is calculated by the Monte Carlo method. We use data sampled from the replay buffer to estimate $p(z_t)$.

\begin{equation}
\begin{aligned} 
p(z_t | s_t, a_t) &=\frac{p(s_t, z_t, a_t)}{p(s_t, a_t)} =\frac{p(a_t | s_t, z_t) p(s_t, z_t)}{p(s_t, a_t)} =\frac{p(a_t | s_t, z_t) p(z_t | s_t) p(s_t)}{p(a_t|s_t) p(s_t)} 
\\ &=\frac{p(a_t | s_t, z_t) p(z_t | s_t)}{\sum_{z^{\prime}} p\left(a_t | s_t, z_t^{\prime}\right) p\left(z_t^{\prime} | s_t\right)},
\end{aligned}
\end{equation}

\begin{equation}
p(z_t)=\int p(s_t,a_t) p(z_t | s_t, a_t) \mathrm{d} a_t \mathrm{d} s_t=E_{(s_t,a_t) \sim \mathcal{D}}[p(z_t | s_t, a_t)],
\end{equation}

where $\mathcal{D}$ is a data batch used while training. We use this method for a more precise estimation. Then we can calculate corresponding entropy, which is estimated by the Monte Carlo method while training.

\begin{equation}
\begin{aligned} H(z_t|s_t,a_t) &=E[-p( z_t| s_t,a_t) \log p(z_t | s_t,a_t)] \end{aligned}
\end{equation}

\begin{equation}
\begin{aligned} H(z_t) &=E[-\log p(z_t)] \end{aligned}
\end{equation}

\section{Experiment Details}

\subsection{Hyperparameters}

In this paper, all hyperparameters of SAC, DAC and adInfoHRL follow the original papers~\cite{haarnoja2018soft,zhang2019dac,osa2019hierarchical}. We directly utilize code uploaded by related authors on github to achieve similar performances as original papers. In SAC, we choose the latest algorithm which updates $\alpha$ while training. In order to get the best comparison effect, we set most hyperparameters same as SAC. Different from SAC, we fix $\alpha$ to one and set the scale of rewards to five for all tasks. It should be noticed that only the agent trained on Walker2d-v2 is greatly affected by random seeds. After rough adjustment, we set the mutual information weight to 0.3 on Walker2d-v2, which is different from other environments. Meanwhile, it should be noticed that we do not use popart~\cite{hessel2019multi} in SAC but use it in SOAC. Actually, we have tested that adding popart to SAC will dramatically weaken its performance. In contrast, popart in SOAC will stabilize training. All corresponding hyperparameters are listed in Table~\ref{soac-table}. 

%Meanwhile, for reviewers who are wondering what is the effect of popart on SAC, we add another experiment in Figure~\ref{fig:add}. While training SAC with popart, we utilize the same neural network structure as SOAC, set reward scale to five and fix alpha to one. It can be noticed that SAC only benefits from popart on HalfCheetah-v2. So we decided to only show SAC without popart in the main body of this paper.

\begin{table}
  \caption{SOAC Hyperparameters}
  \label{soac-table}
  \centering
  \begin{tabular}{lllll}
    \toprule
    Hyperparameters                     & SOAC      &SAC        & DAC     &adInfoHRL    \\
    \midrule
    \emph{Common hyperparameters} &&&&\\
    \quad Optimizer                           & Adam      & Adam      & Adam    & Adam~\cite{kingma2014adam}\\
    \quad Learning rate                       & 3e-4      & 3e-4      & 3e-4    & 1e-3        \\
    \quad Discount weight($\gamma$)           & 0.99      & 0.99      & 0.99    & 0.99        \\
    \quad Replay buffer size                  &$10^6$     &$10^6$     & 2048    &$10^6$       \\
    \quad Optimization batch size             &256        &256        &64       &100          \\
    \quad Number of units in hidden layers    &(256, 256) &(256, 256) &(64, 64) &(400, 300)   \\
    \quad Nonlinearity                        &ReLU       &ReLU       &ReLU     &ReLU,Tanh    \\
    \quad Target update interval              &1          &1          &1        &1            \\
    \quad Gradient steps                      &1          &1          &10       &1            \\
    \quad Target smoothing coefficient($\tau$)&0.005      &0.005      &-        &0.005        \\
    \quad Option number                       &4          &-          &4        &4            \\
    \quad Reward scale                        &5          &1          &1        &1            \\
    \quad Use Popart~\cite{hessel2019multi}   &True       &False      &False    &False        \\
    \midrule
    \emph{SOAC and adInfoHRL} &&&&\\
    \quad Mutual information weight($\lambda_1$)&1        &-          &-        &0.1          \\
    \quad Noise influence weight($\lambda_2$) &5          &-          &-        &0.04         \\
    \quad Action noise $\boldsymbol{\epsilon}_{\boldsymbol{a}}$&0.2&-  &-        &0.2          \\
    \quad State noise $\boldsymbol{\epsilon}_{\boldsymbol{s}}$&1&-     &-        &1            \\
    \midrule
    \emph{SOAC and SAC} &&&&\\
    \quad Random action steps                 &$10^4$     &$10^4$     &-        &-            \\
    \quad Start training steps                &$10^4$     &$10^4$     &-        &-            \\
    \quad Update alpha                        &False      &True       &-        &-            \\
    \midrule
    \emph{adInfoHRL only} &&&&\\
    \quad Size of the on-policy buffer        &-          &-          &-        &5000         \\
    \quad Total batch size for all option policies&-      &-          &-        &400          \\
    \quad Batch size for the option network   &-          &-          &-        &50           \\
    \quad Number of epochs for training the option network&-&-        &-        &40           \\
    \quad Noise clip threshold                &-          &-          &-        &0.5          \\
    \quad Noise for exploration               &-          &-          &-        &0.1          \\
    \midrule
    \emph{DAC only} &&&&\\
    \quad GAE coefficient                     &-          & -         &0.95     & -           \\
    \quad Action probability ratio clip       &-          &-          &0.2      &-            \\
    \midrule
    \emph{SAC only} &&&&\\
    \quad Entropy target                      &-          &-dim($\mathcal{A}$)&-&-            \\
    \bottomrule
  \end{tabular}
\end{table}

%\subsection{Activation of Options}

%Tasks in Mujoco environments are to go in one direction as fast as possible in a flat place. The actions of agents are periodic and high frequency. In order to cut a short action cycle into different parts with different characteristics, our option selection policy presents a high-frequency switching characteristic. In addition, the action spaces corresponding to different options are distinguishable.

%\begin{figure}[H]\centering
%\includegraphics[width=1\textwidth]{./pic/sac.pdf}
%\caption{Learning curves of transferring option selection policy compared with transferring nothing.}
%\label{fig:add}
%\end{figure}

\subsection{Environment Details}

Our experiments are based on Mujoco domains in OpenAI Gym (https://gym.openai.com/). Reward functions used in Section~\ref{exp:1} are the original reward functions. In Section~\ref{exp:2}, we transfer the final instances of SOAC shown in Figure~\ref{fig:experiment1} to opposite tasks. The reward function in each Mujoco task includes three parts: alive bonus, control cost and moving bonus. To build opposite reward functions, we take the opposite of moving bonus and keep other items unchanged. Meanwhile, the dimensions of state space and action space are listed in Table~\ref{env_table}.

\begin{table}
  \caption{state-Action Space Dimention in Mujoco Domains}
  \label{env_table}
  \centering
  \begin{tabular}{lll}
    \toprule
    Environments      & State dimension      &Action dimension      \\
    \midrule
    Hopper-v2 & 11 & 3 \\
    Walker2d-v2 & 17 & 6 \\
    HalfCheetah-v2 & 17 & 6 \\
    Ant-v2 & 111 & 8 \\
    \bottomrule
  \end{tabular}
\end{table}

\end{appendix}

\end{document}